\documentclass[runningheads]{llncs}
\usepackage[T1]{fontenc}
\usepackage{graphicx}
\usepackage{tabularx}
\usepackage{booktabs}
\usepackage{wrapfig}

\usepackage{longtable}
\begin{document}
\title{Enriched BERT Embeddings for Scholarly Publication Classification}
\titlerunning{Enriched BERT Embeddings for Scholarly Publication Classification}
%
\author{Benjamin Wolff\orcidID{0000-0001-9345-8958} \and
Eva Seidlmayer\orcidID{0000-0001-7258-0532} \and
Konrad U. Förstner\orcidID{0000-0002-1481-2996}}
\authorrunning{Wolff et al.}
%
\institute{ZB MED - Information Centre for Life Sciences, Cologne, Germany}
\maketitle  
\begin{abstract}
With the rapid expansion of academic literature and the proliferation of preprints, researchers face growing challenges in manually organizing and labeling large volumes of articles. The NSLP 2024 FoRC Shared Task I addresses this challenge organized as a competition. The goal is to develop a classifier capable of predicting one of 123 predefined classes from the Open Research Knowledge Graph (ORKG) taxonomy of research fields for a given article.
This paper presents our results.

Initially, we enrich the dataset (containing English scholarly articles sourced from ORKG and arXiv), then leverage different pre-trained language Models (PLMs), specifically BERT, and explore their efficacy in transfer learning for this downstream task.
Our experiments encompass feature-based and fine-tuned transfer learning approaches using diverse PLMs, optimized for scientific tasks, including SciBERT, SciNCL, and SPECTER2. We conduct hyperparameter tuning and investigate the impact of data augmentation from bibliographic databases such as OpenAlex, Semantic Scholar, and Crossref.
Our results demonstrate that fine-tuning pre-trained models substantially enhances classification performance, with SPECTER2 emerging as the most accurate model. Moreover, enriching the dataset with additional metadata improves classification outcomes significantly, especially when integrating information from S2AG, OpenAlex and Crossref. Our best-performing approach achieves a weighted F1-score of 0.7415.
Overall, our study contributes to the advancement of reliable automated systems for scholarly publication categorization, offering a potential solution to the laborious manual curation process, thereby facilitating researchers in efficiently locating relevant resources.
\end{abstract}
\section{Introduction and Background}
In academic publishing, we currently see two trends converging: the exponential growth of publications, on average doubling every 14 years \cite{bornmann2021growth}, as well as the increased frequency of publication, due to the growing importance of preprints, which accelerates the publishing process \cite{hoy2020rise}. The vast amount of publications presents a challenge to researchers in distinguishing relevant papers from irrelevant ones concerning a specific research question. Thus, a fine-grained categorization provides researchers with an invaluable tool for finding relevant resources.
Manual curation is not a feasible approach. Besides that, categorization is not static. Instead, it has a dynamic nature: New categories emerge over time, while others evolve or disappear. Moreover, the compiled categories depend on the underlying platforms, where the publications are indexed. Thus we need reliable automated systems for the task of scholarly publication categorization.
This paper stands within the scope of the NSLP2024\footnote{https://nfdi4ds.github.io/nslp2024/} and is dedicated to the Field of Research Classification (FoRC) shared task, Subtask I.
The goal of Subtask I is to develop a classifier that predicts one of 123 predefined classes from Open Research Knowledge Graph (ORKG) taxonomy of research fields for a given article. The competition hosts provide a dataset of English articles (split into test and validation sets), compiled from ORKG\footnote{https://orkg.org/} and arXiv\footnote{https://arxiv.org/}, which suggests a classifier that is trained in a supervised manner. The dataset consists of a selection of metadata fields. The shared task is run as a competition. Submitted systems will be evaluated on a test set, using accuracy and weighted scores of recall, precision, and F1.

With the advent of transformer models, the field of Natural Language Processing (NLP) underwent significant changes. Pre-trained Language Models are the current state-of-the-art across many NLP tasks and achieve top positions in General Language Understanding Tasks\footnote{https://gluebenchmark.com/}. FoRC Shared Task I represents a task of scholarly document classification, a task where BERT \cite{devlin2018bert}, an encoder model for text embeddings, has demonstrated excellent results in a variety of benchmarks \cite{singh2022scirepeval}. BERT is available in different flavors, pre-trained on specific tasks, and can be further fine-tuned to adapt it to downstream tasks.

Significant research effort is put towards automated classification into scholarly publication categorization. Automating this task could replace the time-consuming and costly manual curation process, which relies on experts in the field. Garcia et al. \cite{garcia-silva} explore how BERT and its variants effectively use specific words, semantically relevant to research fields in their final layer.
In SciRepEval \cite{singh2022scirepeval} the authors introduce a benchmark, which applies 24 realistic tasks for use cases involving scientific document embeddings. The results yield a novel model called SPECTER2.
Ostendorff et al. \cite{ostendorff2019enriching}  show how BERT embeddings, when combined with metadata and knowledge graph embeddings, improve results in a document classification task.

\section{Dataset}
The provided dataset contains the following fields: abstract, author, DOI, URL, publication month, publication year, title, publisher and label. However, these fields are incomplete for many items. Particularly, the absence of a DOI becomes problematic later on during the enrichment process, as we can not use a unique identifier for further information retrieval from bibliometric databases. In total, the dataset comprises 50,441 items (train: 41,540, validation: 8901). For evaluation purposes, initially, a test set without labels was provided, which was distributed with labels after the end of the competition's deadline. 
The dataset (train and validation) shows significant class imbalance with a multitude of items assigned to the fields ''Physics and Mathematics'' ($\approx$ 79.5\%) , while other fields like ''Arts and Humanities'' are largely underrepresented, with only a handful of entries ($\approx$ 0.14\%).

\section{Enrichment}\label{sec:enrichment}
We encountered a challenge with enrichment due to the absence of DOIs in the provided article metadata collection. In the training dataset, 16,977 DOIs were missing, while the test and validation datasets had 3,712 and 3,673 gaps in the DOI column, respectively. To fill the missing DOI, we used the article titles to query the OpenAlex(OA)\footnote{https://openalex.org/} database, a bibliographic repository containing research literature metadata. In cases where the title was not suitable for retrieval, we decided to clean it by removing special characters like line breaks, tabs, multiple spaces, as well as LaTeX tags. Subsequently, we successfully retrieved DOIs for numerous titles, filling 7,921 gaps in the training dataset and adding 1,702 and 1,710 DOIs to the test and validation datasets, respectively.

For further metadata enrichment we used three bibliographic databases: OpenAlex, Semantic Scholar Academic Graph(S2AG)\footnote{https://www.semanticscholar.org/product/api} and  Crossref(CR)\footnote{https://www.crossref.org/}. These sources were employed to fill gaps in the provided original dataset and enhance it with additional categories. S2AG is a large, open, heterogeneous knowledge graph of scholarly works, authors and citations, developed by Allen Institute for AI\footnote{https://allenai.org/}, suitable for enrichment purposes. CR is an association of publishers that provides a database of metadata for scientific literature.
From the mentioned providers, we collected the following metadata:
\begin{itemize}
    \item Openalex: topics, subtopics, concepts, keywords and external identifiers
    \item S2AG: fields of study
    \item Crossref: journal title, (research) subjects
\end{itemize}
\section{Approaches}

\subsection{BERT-Embeddings}

In our first approach we decided to use a single BERT model to derive document embeddings, followed by a dense layer for the classification task. Since the BERTbase model uses 12 transformer blocks, and a hidden dimension size of 768, a single dense layer is sufficient to separate the documents for classification. To fully leverage the power of pre-trained models and achieve a good performance, there are two important questions to consider:
\begin{enumerate}
    \item Which available pre-trained models suit the given task?
    \item Which hyperparameter setting yields the best performance?
\end{enumerate}
Furthermore, we aim to investigate the impact of transfer learning on selected pre-trained models by comparing feature-based transfer learning and fine-tuning.  Additionally, we examine the effect of enriched data on the classification result.

\subsubsection{Selecting Pre-trained Models}

For selecting promising pre-trained models for our downstream task, we rely on the insights from the SciRepEval benchmark \cite{singh2022scirepeval} and chose the top-performing models identified for scientific tasks as potential candidates: SciBERT \cite{beltagy2019scibert}, SciNCL \cite{ostendorff2022neighborhood} and SPECTER2 \cite{singh2022scirepeval}. Additionally, the BERTbase model served as a baseline \cite{devlin2018bert}.
SciBERT, a variant of BERT, is specifically trained on a vast collection of scientific publications to optimize its performance on scientific downstream tasks. SciNCL, initialized with the SciBERTs weights, utilizes citation graph neighborhood to generate samples for Contrastive Learning, a technique used to learn representations by contrasting positive and negative examples.
SPECTER2 employs a BERTbase model, which is trained from scratch using citation links, similar to the original SPECTER model. However, the training data for SPECTER2 comprises 10 times more triplets spanning 23 Fields of Study\footnote{https://github.com/allenai/SPECTER2}.

\subsubsection{Hyperparameter Tuning}
To find the best hyperparameter settings, we performed a grid search. In grid search we define a range of values for each hyperparameter. All possible combinations of these values are explored systematically and evaluated to identify the best-performing combination. The models' performance was evaluated on BERTbase with weighted F1-Score on the test set. We employed the following parameters and ranges for grid search (best parameter combination is highlighted): Batchsize:  [8, 16, \textbf{32}], Learningrate: [$1\mathrm{e{-3}}$, $\mathbf{1e{-4}}$, $1\mathrm{e{-5}}$], Weight Decay: [$1\mathrm{e}{-1}$, $1\mathrm{e}{-2}$, $\mathbf{1e{-3}}$].
We experimented with a range of 3-5 training epochs. During evaluation, we observed signs of overfitting after 3 epochs. This was evident by a sharp decrease in loss on the training data while loss on the validation set simultaneously increased. We maintained hyperparameter settings constant for all experiments.

\subsubsection{Transfer Learning}
For an assessment of the selected pre-trained models, we evaluated the performance between the feature-based and fine-tuned transfer learning:
In feature-based transfer learning parameters of the pre-trained models were frozen. Only the task-specific classifier on top was trained, in the fine-tuned transfer learning the parameters of the pre-trained models are additionally adapted to the task. For the basic evaluation, we did not utilize enrichments. We relied solely on title and abstract. For a more detailed overview, we ran both approaches on the models selected in the previous step. 

\newcolumntype{Y}{>{\centering\arraybackslash}X}

\begin{table}[htb]
\begin{tabularx}{\textwidth}{X cccc cccc}
\toprule
 & \multicolumn{4}{c}{Feature-Based} & \multicolumn{4}{c}{Finetuned} \\
\cmidrule(lr){2-5} \cmidrule(lr){6-9}
\multicolumn{1}{l}{} & \multicolumn{1}{Y}{Acc}  & \multicolumn{1}{Y}{Prec} & \multicolumn{1}{Y}{Rec} & \multicolumn{1}{Y}{F1} & \multicolumn{1}{Y}{Acc} & \multicolumn{1}{Y}{Prec} & \multicolumn{1}{Y}{Rec} & \multicolumn{1}{Y}{F1} \\ 
\multicolumn{1}{l}{BERTbase} & \multicolumn{1}{Y}{0.2033}    & \multicolumn{1}{Y}{0.0887}     & \multicolumn{1}{Y}{0.2033}    & \multicolumn{1}{Y}{0.1013} & \multicolumn{1}{Y}{0.6973} & \multicolumn{1}{Y}{0.6838}  & \multicolumn{1}{Y}{0.6973}  & \multicolumn{1}{Y}{0.6868} \\ 
\multicolumn{1}{l}{SciBERT}  & \multicolumn{1}{Y}{0.3313}    & \multicolumn{1}{Y}{0.2368}     & \multicolumn{1}{Y}{0.3313}    & \multicolumn{1}{Y}{0.2179} & \multicolumn{1}{Y}{0.7259} & \multicolumn{1}{Y}{0.7223}  & \multicolumn{1}{Y}{0.7259}  & \multicolumn{1}{Y}{0.7206} \\ 
\multicolumn{1}{l}{SciNCL} & \multicolumn{1}{Y}{\underline{0.490}}    & \multicolumn{1}{Y}{\underline{0.4497}}     & \multicolumn{1}{Y}{\underline{0.490}} & \multicolumn{1}{Y}{\underline{0.4095}} & \multicolumn{1}{Y}{0.7285} & \multicolumn{1}{Y}{0.7245}  & \multicolumn{1}{Y}{0.7285}  & \multicolumn{1}{Y}{0.7239}  \\ 
\multicolumn{1}{l}{SPECTER2} & \multicolumn{1}{Y}{0.4351}    & \multicolumn{1}{Y}{0.4035}     & \multicolumn{1}{Y}{0.4351}    & \multicolumn{1}{Y}{0.3364} & \multicolumn{1}{Y}{\underline{0.7330}} & \multicolumn{1}{Y}{\underline{0.7285}}  & \multicolumn{1}{Y}{\underline{0.7330}}  & \multicolumn{1}{Y}{\underline{0.7283}} \\
\bottomrule
\end{tabularx}
\caption{Comparison of evaluation metrics for feature-based and finetuned transfer learning on a selection of different PLMs.}
\label{table:feature_vs_finetuned}
\end{table}

\subsubsection{Raw Data vs Enriched Data}
\begin{figure}[ht]
\includegraphics[width=\textwidth]{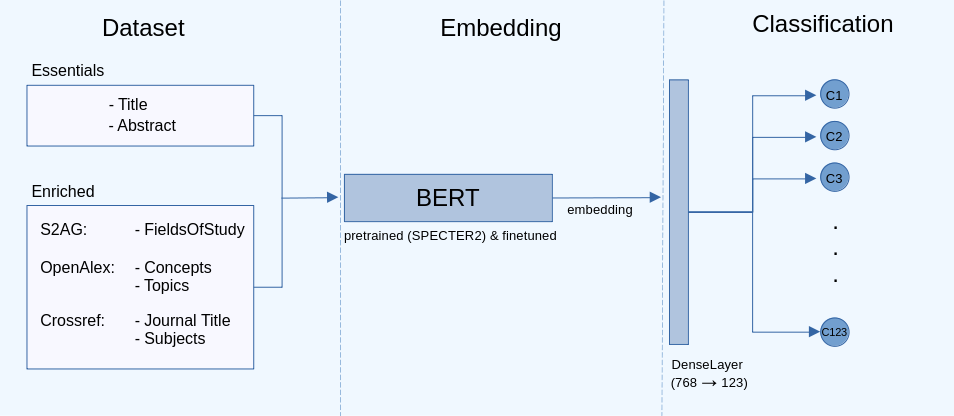}
\caption{Workflow using a single BERT-model with enrichments and SPECTER2 (best performing model according to Table \ref{table:enrichments})} \label{fig:bert_enriched}
\end{figure}

We utilized the gathered enrichments to enhance the classification performance, assuming that these additional pieces of information could significantly influence the outcome. Since BERT has a limited input capacity (maximum of 512 tokens), we decided to use only a selection of the data enrichments. Moreover, it is crucial to determine the order and manner in which the enrichments are fed into the model. First, we selected enrichments from which we expected the most informative content, while consuming few additional input tokens. We first experimented with the data gathered from S2AG and OpenAlex and subsequently integrated  enrichments from CrossRef in a second step. We decided to integrate the following enrichments:
\begin{itemize}
    \item from S2AG: Fields of Study (FoS)
    \item from OpenAlex: Concepts and Topics
    \item from CrossRef: Journal Title, Subjects
\end{itemize}
We opted for the following order: title, fields of study, topics, abstract, concepts, categories, journal title and subject. Each of the fields was separated by a [SEP]-token. Additionally, the field name was prepended to the terms to provide context (except for title and abstract). Our final pipeline is depicted in Figure \ref{fig:bert_enriched}.

\begin{table}[htb]
\begin{tabularx}{\textwidth}{X cccc cccc}
\toprule
 & \multicolumn{4}{c}{SPECTER2} & \multicolumn{4}{c}{SciNCL} \\
\cmidrule(lr){2-5} \cmidrule(lr){6-9}
\multicolumn{1}{l}{} & \multicolumn{1}{Y}{Acc}  & \multicolumn{1}{Y}{Prec} & \multicolumn{1}{Y}{Rec} & \multicolumn{1}{Y}{F1} & \multicolumn{1}{Y}{Acc} & \multicolumn{1}{Y}{Prec} & \multicolumn{1}{Y}{Rec} & \multicolumn{1}{Y}{F1} \\ 
\multicolumn{1}{l}{Title+Abstract(TA)} & \multicolumn{1}{Y}{0.7330} & \multicolumn{1}{Y}{0.7285}  & \multicolumn{1}{Y}{0.7330}  & \multicolumn{1}{Y}{0.7283} & \multicolumn{1}{Y}{0.7285} & \multicolumn{1}{Y}{0.7245}  & \multicolumn{1}{Y}{0.7285}  & \multicolumn{1}{Y}{0.7239} \\ 

\multicolumn{1}{l}{TA+S2AG+OA} & \multicolumn{1}{Y}{0.7419} & \multicolumn{1}{Y}{0.7386} & \multicolumn{1}{Y}{0.7419}  & \multicolumn{1}{Y}{0.7367} & \multicolumn{1}{Y}{0.7353} & \multicolumn{1}{Y}{0.7299}  & \multicolumn{1}{Y}{0.7353}  & \multicolumn{1}{Y}{0.7256} \\

\multicolumn{1}{l}{TA+S2AG+OA+CR} & \multicolumn{1}{Y}{\underline{0.7467}} & \multicolumn{1}{Y}{\underline{0.7438}}  & \multicolumn{1}{Y}{\underline{0.7467}}  & \multicolumn{1}{Y}{\underline{0.7415}} & \multicolumn{1}{Y}{0.7340} & \multicolumn{1}{Y}{0.7275}  & \multicolumn{1}{Y}{0.7340}  & \multicolumn{1}{Y}{0.7246} \\
\bottomrule
\end{tabularx}
\caption{Comparison of evaluation metrics for raw data (title and abstract) and enriched data from S2AG, OpenAlex(OA) and CrossRef (CR).}
\label{table:enrichments}
\end{table}

\subsection{Combined BERT-Embeddings (TwinBERT)}
\begin{figure}[ht]
\includegraphics[width=\textwidth]{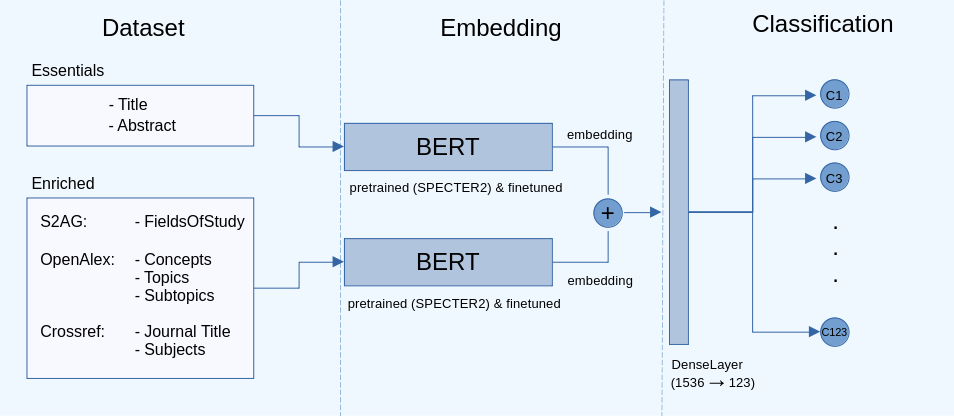}
\caption{Twin-Bert Model, which utilizes two BERT models, to separate embeddings from essential data (title, abstract) from enriched data.} \label{fig:twinbert}
\end{figure}

We explored alternative approaches to enhance the methodology described earlier, considering BERT limitations: 512 token limit and treating the entire input as a single document.
To overcome these issues, we experimented with a custom model that employs two BERT models in parallel. In literature, we found several similar approaches already known as TwinBERT(\cite{lu2020twinbert}, \cite{gombert2021twin}), but in different contexts. The use of two BERT models provides separate embedding pathways: one for the basic document (title and abstract) and another for metadata. These embeddings are then concatenated before classification.
By utilizing two BERT models, the number of parameters more than doubles. Thus we aimed to determine if this additional effort is justified. When experimenting with TwinBERT, we varied the number of transformer layers, to account for this consideration. We developed TwinBERT as a custom model using PyTorch \cite{paszke2019pytorch}. Our TwinBERT model is depicted in Figure \ref{fig:twinbert}. The results are presented in Table \ref{table:twinbert}.

\begin{table}[htb]
\begin{tabularx}{\textwidth}{X cccc cccc}
\toprule
 & \multicolumn{4}{c}{TwinBERT} \\
\cmidrule(lr){2-5}
\multicolumn{1}{l}{\textbf{\# of Layers}} & \multicolumn{1}{Y}{Acc}  & \multicolumn{1}{Y}{Prec} & \multicolumn{1}{Y}{Rec} & \multicolumn{1}{Y}{F1} \\ 
\multicolumn{1}{l}{12 Layers} & \multicolumn{1}{Y}{0.1145} & \multicolumn{1}{Y}{0.0131}  & \multicolumn{1}{Y}{0.1145}  & \multicolumn{1}{Y}{0.0235}  \\ 
\multicolumn{1}{l}{8 Layers} & \multicolumn{1}{Y}{0.1280} & \multicolumn{1}{Y}{0.0278}  & \multicolumn{1}{Y}{0.1280}  & \multicolumn{1}{Y}{0.0444}  \\ 
\multicolumn{1}{l}{6 Layers} & \multicolumn{1}{Y}{0.6179} & \multicolumn{1}{Y}{0.5940}  & \multicolumn{1}{Y}{0.6179}  & \multicolumn{1}{Y}{0.5959}  \\ 
\multicolumn{1}{l}{4 Layers} & \multicolumn{1}{Y}{0.6688} & \multicolumn{1}{Y}{0.6560}  & \multicolumn{1}{Y}{0.6688}  & \multicolumn{1}{Y}{0.6577}  \\ 
\multicolumn{1}{l}{2 Layers} & \multicolumn{1}{Y}{\underline{0.6719}} & \multicolumn{1}{Y}{\underline{0.6597}}  & \multicolumn{1}{Y}{\underline{0.6719}}  & \multicolumn{1}{Y}{\underline{0.6618}}  \\ 
\bottomrule
\end{tabularx}
\caption{Comparison of evaluation metrics for TwinBERT model with varying number of transformer layers.}
\label{table:twinbert}
\end{table}

\section{Results}
Table \ref{table:feature_vs_finetuned} shows the evaluation results for transfer learning without any enrichments.  In feature-based transfer learning (no fine-tuning) across different models, we achieve a weighted F1-score ranging from 0.1013 to 0.4095.  While this result demonstrates a general ability to address the required task, the quality falls short of an acceptable range. Nevertheless, models optimized for scientific publications outperform the BERTbase model with a significant margin, with SciNCL leading. In fine-tuned transfer learning, the results for these models are closely clustered, with the weighted F1-score ranging from 0.6868 to 0.7283. 
SciBERT, SciNCL, and SPECTER2 achieve F1-scores that are nearly equivalent (between 0.7206 and 0.7283), with SPECTER2 now taking the lead position. These results suggest that for the given task, fine-tuning plays a more significant role than the actual selection of the pre-trained model. Compared to the BERTbase model, the models pre-trained for scientific tasks perform significantly better, although the results of the fine-tuned  BERT Base model are quite acceptable.

Table \ref{table:enrichments} presents the evaluation results for finetuned transfer learning, incorporating enrichments. We decided to assess the two best-performing pre-trained models from the preceding task (SciNCL and SPECTER2). With enrichments, we observe an improvement in the evaluation metrics for both models. However the enrichments yield a more substantial improvement for SPECTER2. The F1-score for SPECTER2 increases by 0.013, while for SciNCL, it only rises by a slight margin of 0.0017. We find that combining enrichments from multiple sources (S2AG, OA, CR) enhances the classification process without adding noise.
Our evaluation demonstrates the superior performance of the SPECTER2 model, fine-tuned for the task. The best results are achieved with enrichments from various sources (S2AG, OA, CR) with an F1-score of 0.7415.

Table \ref{table:twinbert} shows the evaluation results for our custom TwinBERT model. Despite its increased complexity, and the option to incorporate additional enrichments, this model yields unsatisfactory results. 
During the evaluation, we simplified the model by gradually reducing the number of transformer layers. With each layer removed, the model's performance improved. The best performance was achieved with only 2 layers in each BERT Model. However, with an F1-score of 0.6618, it still lags behind the simplest BERTbase model without any enrichments.

\section{Discussion \& Conclusion}
We present a classification model for scholarly publications that achieves a weighted F1-Score of 0.7415 in the FoRC shared task I. When considering the evaluation results, it is noteworthy that even with a fine-tuned BERTbase model without any enrichments, a remarkable weighted F1 of 0.6868 is achieved. This demonstrates the robustness and versatility of the BERTbase model.
However, the models optimized for scholarly publications perform noticeably better, yet these models are almost equivalent without enrichments. By enriching from various sources, we were able to further boost the classification performance, with SPECTER2 showing a better response to these enrichments compared to other models. To appropriately contextualize and evaluate our results, a comparison with human curators would be valuable.
The provided dataset represents reality in a highly simplified manner. In our dataset, there is only one target label per item. Thus, in some cases, there are numerous alternative categorizations that deviate from the provided solution but are nonetheless not incorrect due to thematically overlapping categories.  Instead of providing a single correct label, training the model with multiple possible classes per item could lead to a more nuanced prediction. Additionally the dataset is highly unbalanced, with a strong emphasis on "Physical Sciences and Mathematics", leading to a bias in the model. Training with hundreds or thousands of samples greatly enhances class efficacy compared to having only a few (e.g., "Art and Humanities"), which limits the transferability of our trained model.
The approach of a two-stage TwinBERT, based on theoretical considerations, failed to deliver significant advantages in practical application.
Although we applied the same methodology to compute the evaluation results, we were unable to reproduce the competition's outcomes, leading to a weighted F1-score approximately 0.008 lower.

Our current approach might potentially achieve even better results by incorporating a dedicated SPECTER2 adapter specifically designed for classification tasks.
We also intend to explore a sliding window approach, addressing the input limitations of BERT(\cite{pappagari_hierarchical_2019}, \cite{wang-etal-2018-multi-granularity}), to enable the model to leverage the full scope of enriched metadata and full-text.  Our source code, as well as the trained models are available for reuse\footnote{https://github.com/foerstner-lab/NSLP2024-FoRC} under the MIT-License.

\bibliographystyle{splncs04}
\bibliography{references.bib}

\end{document}